\documentclass[10pt,twocolumn,letterpaper]{article}

\usepackage[pagenumbers]{cvpr} % To force page numbers, e.g. for an arXiv version

\usepackage{graphicx}
\usepackage{amsmath}
\usepackage{amssymb}
\usepackage{booktabs}

\usepackage{caption}
\usepackage{subcaption}
\usepackage[dvipsnames]{xcolor}

\newif\ifarxiv
\arxivtrue

\usepackage[pagebackref,breaklinks,colorlinks]{hyperref}

\usepackage[capitalize]{cleveref}
\crefname{section}{Sec.}{Secs.}
\Crefname{section}{Section}{Sections}
\Crefname{table}{Table}{Tables}
\crefname{table}{Tab.}{Tabs.}

\def\cvprPaperID{7} % *** Enter the CVPR Paper ID here
\def\confName{CVPR}
\def\confYear{2022}

\begin{document}

\title{Fix the Noise: Disentangling Source Feature for Transfer Learning of StyleGAN}

\author{Dongyeun Lee \quad Jae Young Lee \quad Doyeon Kim \quad Jaehyun Choi \quad Junmo Kim \\
KAIST \\
{\tt\small \{ledoye, mcneato, doyeon\_kim, chlwogus, junmo.kim\}@kaist.ac.kr}
% \and
}

\twocolumn[{%
\renewcommand\twocolumn[1][]{#1}%
\maketitle
\vspace{-7mm}
\begin{center}
    \centering
    \begin{tabular}{c}
        \begin{minipage}{0.9\textwidth}
        \includegraphics[width=\linewidth]{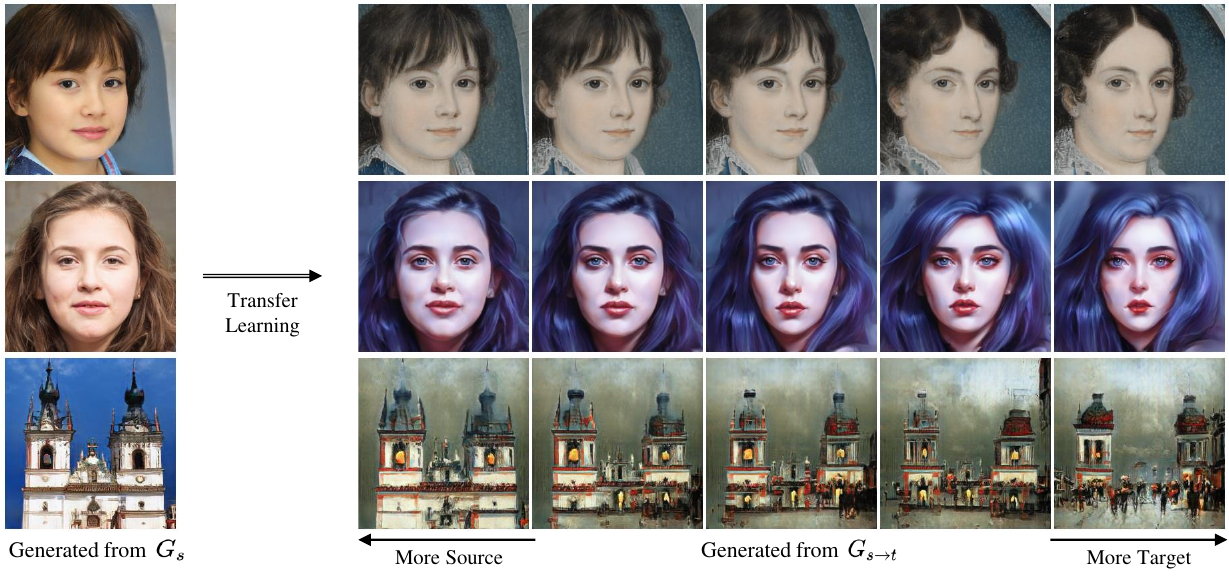}
        \end{minipage}
\hspace{-0.4cm}
\vspace{-0.1cm}
\\

\end{tabular}
\vspace{-0.2cm}
\captionof{figure}{Given a source model $G_s$, our method can smoothly control the degree of source domain features in a fine-tuned model $G_{s \rightarrow t}$. 
Images in each row are generated from the same latent code $\mathbf z \in \mathcal Z$ by $G_s$ and $G_{s \rightarrow t}$. 
}
\vspace{-0.0cm}
\end{center}%
}]

\maketitle

%%%%%%%%% ABSTRACT
\begin{abstract}
\vspace{-3mm}
   Transfer learning of StyleGAN has recently shown great potential to solve diverse tasks, especially in domain translation.
   Previous methods utilized a source model by swapping or freezing weights during transfer learning, however, they have limitations on visual quality and controlling source features.
   In other words, they require additional models that are computationally demanding and have restricted control steps that prevent a smooth transition.
   In this paper, we propose a new approach to overcome these limitations.
   Instead of swapping or freezing, we introduce a simple feature matching loss to improve generation quality.
   In addition, to control the degree of source features, we train a target model with the proposed strategy, FixNoise, to preserve the source features only in a disentangled subspace of a target feature space.
   Owing to the disentangled feature space, our method can smoothly control the degree of the source features in a single model.
   Extensive experiments demonstrate that the proposed method can generate more consistent and realistic images than previous works. 
   
\end{abstract}

%%%%%%%%% BODY TEXT
\section{Introduction}
\label{sec:intro}
Generative adversarial networks (GANs) have recently made tremendous strides in the field of high-quality image synthesis, especially using style-based architecture \cite{karras2019style,karras2020analyzing,Karras2021}.
Many studies have utilized pre-trained StyleGAN \cite{karras2019style,karras2020analyzing} in various fields, such as image encoding \cite{zhu2020indomain,xu2021generative,richardson2021encoding}, discovering latent semantics \cite{harkonen2020ganspace,shen2021closed,wu2021stylespace}, image editing \cite{viazovetskyi2020stylegan2,collins2020editing} and transfer learning \cite{Karras2020ada,mo2020freeze,ojha2021few}.
In particular, StyleAlign \cite{wu2022stylealign} analyzed transfer learning of StyleGAN2 \cite{karras2020analyzing} and revealed several relationships between a pre-trained source model and a target model that is fine-tuned on a target domain.

In StyleAlign, it was observed that $\mathcal{W}$ spaces of the source and target models are similar and these spaces could be viewed as a single shared latent space. 
This observation is parallel with assumptions of several domain translation works \cite{liu2017unsupervised,huang2018multimodal,liu2019few}.
Owing to this characteristic, there have been many attempts to integrate the source and target models into the domain translation task by mapping images to latent space \cite{pinkney2020resolution,lee2020freezeg,kwong2021unsupervised}.
These studies implemented domain translation by embedding an image from the source domain to the latent space of the source model and providing the obtained latent code into the target model to generate a target domain image.
Since it is important to preserve semantic features 
of source images in the domain translation, these studies used several techniques to preserve semantic features during transfer learning.
Layer-swap \cite{pinkney2020resolution} generated a target domain image with coarse spatial characteristics of the source domain by combining low resolution layers of a source generator and high resolution layers of a target generator.
By adjusting the number of layers to be swapped, their method can control the degree of remaining source features.
Freeze G \cite{lee2020freezeg} obtained a similar effect by freezing weights of initial layers of the generator during transfer learning.
UI2I StyleGAN2 \cite{kwong2021unsupervised} froze mapping layers to ensure the same $\mathcal{W}$ space between the source and target models, and combined it with Layer-swap.

Previous works have shown satisfying results, however, there remain several limitations in terms of visual quality and controllability.
First, due to the discrepancy between the two domains, the target model could produce unrealistic images when simply freezing or swapping weights.
For example, we observe that source domain color remains unnaturally in the results 
of previous approaches.
Second, to control the degree of preserved source features, previous works require additional models and their control levels are restricted to the number of layers.
To be specific, previous works need to convert weights from the source model \cite{pinkney2020resolution,kwong2021unsupervised} or train new models \cite{lee2020freezeg} to control the degree, which increase computational cost.
In addition, we observe several inconsistent transitions (\textit{e.g.}, changes in background or the human identity) while controlling the source features.

In this paper, we introduce a new training strategy, FixNoise, to overcome these limitations for preserving semantic features of the source during transfer learning.
To control the degree of source features in a single model, the source and target features must be disentangled in a feature space of the target model.
Our main idea is to preserve the source features only to a particular subset of the target feature space.
Instead of freezing or swapping which generates unnatural results, we apply a simple feature matching loss between the source and target models only in a subspace of the target feature space by fixing noise added after each convolution.
By linear interpolation between the fixed and random noise, our method can smoothly control the source features in a single model without limited control steps.

\section{Method}
\label{sec:method}
Given a source domain model $G_s$, our goal is to train a target domain model $G_{s \rightarrow t}$ initialized with the source model weights while preserving source domain features.
We first briefly discuss the space in which to preserve features and introduce a simple but effective feature matching loss.
Then we propose FixNoise that ensures disentanglement between the two domains features in the feature space of $G_{s \rightarrow t}$. 

\subsection{Which feature to preserve?}
Remark that StyleGAN2 \cite{karras2020analyzing} contains two types of feature spaces: an intermediate feature space that consists of feature convolution layer outputs and a RGB space that consists of RGB outputs transformed from an intermediate feature by tRGB layers.
We refer to the intermediate feature space as $\mathcal H$.
Between the two spaces in StyleGAN2, we choose to preserve the intermediate feature space for the following reasons.
First, it has recently been found that the feature convolution layers 
change the most among layers during transfer learning \cite{wu2022stylealign}.
This observation indicates that the source features mostly vanish in $\mathcal H$.
Second, matching features of the source and target models in $\mathcal H$ enables the subsequent tRGB layers to learn the target distribution.

From the same latent code $\mathbf z \in \mathcal Z$, we encouraged the target model $G_{s \rightarrow t}$ to have similar features as those of the source model $G_s$ in $\mathcal H$ using a simple feature matching loss:

\vspace{-2mm}
\begin{equation}
    \mathcal L_{fm} = \frac{1}{L}\sum_{l=0}^L \big(G_s^l(\mathbf z) -G_{s \rightarrow t}^l(\mathbf z)\big)^2,
\end{equation}
where $L$ denotes the number of feature convolution layers.
Recall that losses that utilize the intermediate features of a network are widely used in GANs literature, such as perceptual loss \cite{johnson2016perceptual,zhang2018perceptual}.
However, the main difference between $\mathcal L_{fm}$ and perceptual loss is in which space the features are matched.
Our loss encourages the source and target models to have similar intermediate features internally, whereas the perceptual loss \cite{zhang2018perceptual} encourages it in the feature space of the external network which is unrelated to image generation.
Therefore, with the proposed loss, we can encourage $G_{s \rightarrow t}$ to have a shared feature space with $G_s$ internally. 

\subsection{Disentangled feature space using FixNoise}

\begin{figure}[t]
    \centering
    \begin{tabular}{ccc}
    \begin{minipage}{0.19\textwidth}\includegraphics[trim=0cm 0cm 0cm 0cm,clip,width=\linewidth]{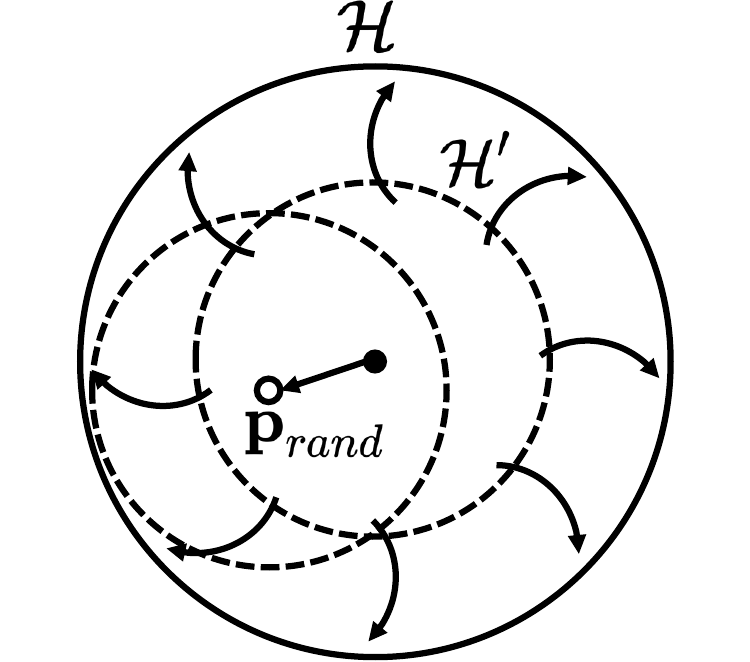}
    \end{minipage} &
    \begin{minipage}{0.19\textwidth}\includegraphics[trim=0cm 0cm 0cm 0cm,clip,width=\linewidth]{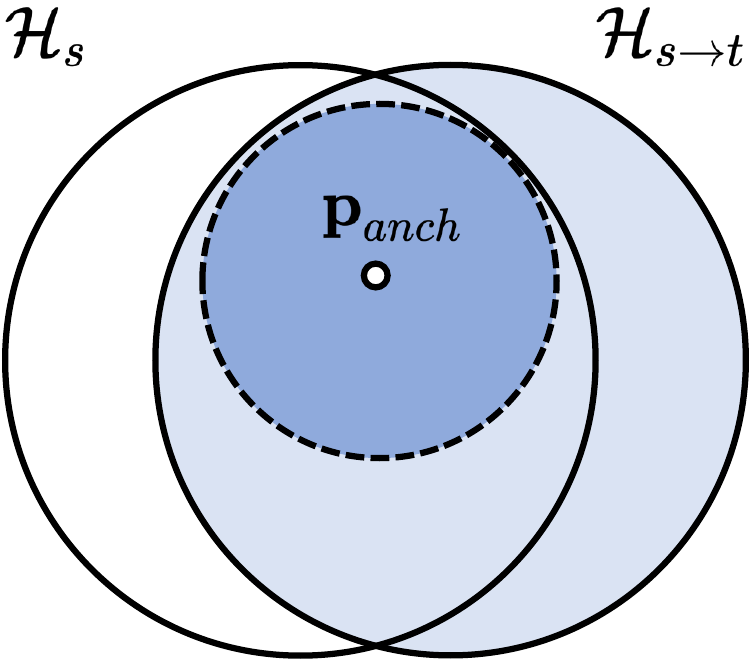}
    \end{minipage}
    \vspace{2mm}
\\
 
(a) & 
(b)  \\

\end{tabular}
\vspace{-2mm}
\caption{An illustration of FixNoise. 
(a) The black dot indicates $\mathbf 0$ noise corresponding to $\mathcal H'$. 
 Randomly sampled noise expands $\mathcal H'$ to $\mathcal H$.
(b) Anchored subspace is denoted by a dotted line. Source features are only mapped to the anchored subspace of $\mathcal H_{s \rightarrow t}$.}
\vspace{-5mm}
\label{fig:2}
\end{figure}

\begin{figure*}[ht!]
\centering
\includegraphics[width=1\linewidth]{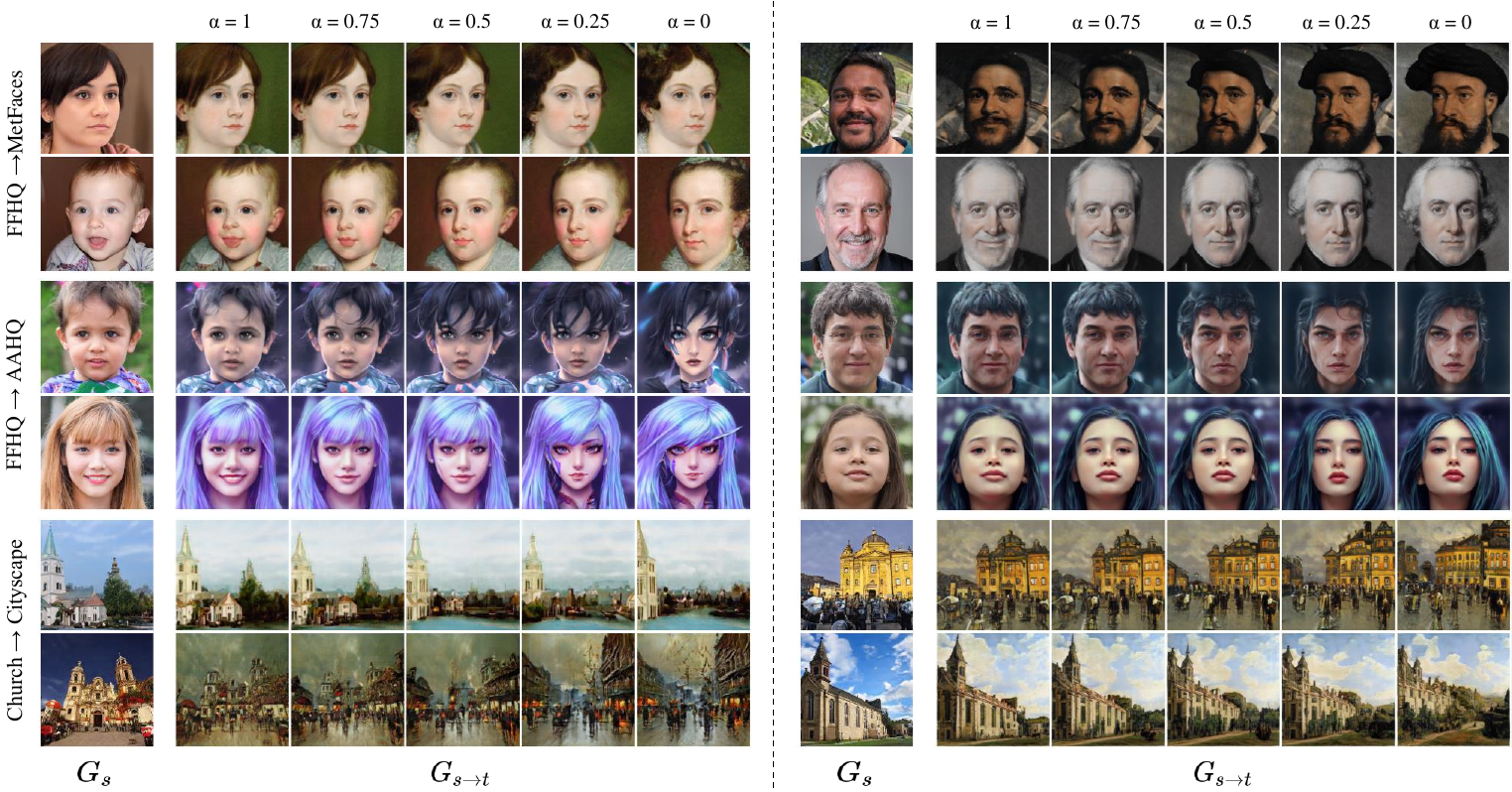}
   \vspace{-5mm}
   \caption{Noise interpolation results on different settings. The interpolation weight $\alpha$ is presented above each column.}
   \vspace{-5mm}
\label{fig:blending}
\end{figure*}

In the optimal case, the loss $\mathcal L_{fm}$ enforces the entire feature space of the target model $\mathcal H_{s \rightarrow t}$ to be the same as that of the source model.
This may disturb $G_{s \rightarrow t}$ to learn diverse target features that do not exist in the source domain.
Even if the target features are learned, the degree of the source features cannot be controlled if the source and target features are entangled in $\mathcal H_{s \rightarrow t}$.
Instead of applying the loss $\mathcal L_{fm}$ to the entire space $\mathcal H_{s \rightarrow t}$, we introduce an effective strategy, FixNoise, that does not disturb target feature learning and allows the different domain features to be disentangled from each other in $\mathcal H_{s \rightarrow t}$.
Our method begins with an assumption that both can be achieved if the source features are mapped in a particular subspace in $\mathcal H_{s \rightarrow t}$.
\vspace{-1mm}

In StyleGAN2, per pixel Gaussian noise is added after each convolution which creates stochastic variation on the generated images.
A feature $\mathbf h' \in \mathcal H'$ is deterministically generated from a latent code, where $\mathcal H'$ is a subspace of $\mathcal H$ with noise $= \mathbf 0$.
As depicted in Fig.~\ref{fig:2}, when each randomly sampled noise $\mathbf p_{rand}$ is added to $\mathbf h'$, it shifts $\mathcal H'$ to a space that corresponds to each $\mathbf p_{rand}$, and consequently expands $\mathcal H'$ to $\mathcal H$.
It signifies the feature space $\mathcal H$ consists of subspaces corresponding to each random noise.
To ensure that source features are only mapped to a particular subspace of $\mathcal H_{s \rightarrow t}$, we fix the noise to a single predefined value when 
$\mathcal L_{fm}$ is applied.
We refer to the fixed noise as an \textit{anchor point} $\mathbf p_{anch}$ and the corresponding subspace as an \textit{anchored subspace}.
Through this, the source features are only mapped in the anchored subspace, while target features are freely adapted to the entire $\mathcal H_{s \rightarrow t}$.
We believe that common features of the two domains are embedded in the anchored subspace, and features that exist only in the target are embedded in the remainder space of $\mathcal H_{s \rightarrow t}$.
By the anchor point, the disentanglement between the different domain features can be achieved in 
$\mathcal H_{s \rightarrow t}$.
This enables smooth transition between images by linear interpolation of the anchor point and random noise, $\alpha \cdot \mathbf p_{anch} + (1-\alpha) \cdot \mathbf p_{rand}$, where $\alpha$ represents the interpolation weight.

\vspace{-1mm}
\section{Experiments}
\vspace{-1mm}
We conduct experiments for several settings considering spatial similarity between the source and target domains.
For similar domain setting, we transfer FFHQ \cite{karras2019style} to MetFaces \cite{Karras2020ada} and AAHQ \cite{liu2021blendgan}.
For distant domain setting, we transfer LSUN Church \cite{yu2015lsun} to WikiArt Cityscape \cite{wikiart}.
All experiments are conducted on $256 \times 256$ resolution images.

\setlength{\tabcolsep}{4pt}
\begin{table}[t!]
\begin{center}
\begin{tabular}{|c|cc|cc|cc|}
\hline
Source        & \multicolumn{4}{c|}{FFHQ}                                  & \multicolumn{2}{c|}{Church}                \\ \hline
Target        & \multicolumn{2}{c|}{MetFaces}  & \multicolumn{2}{c|}{AAHQ} & \multicolumn{2}{c|}{Cityscape}             \\ \hline
$\alpha$& FID            & LPIPS            & FID           & LPIPS             & FID               & LPIPS             \\ \hline
1       & 40.37          & \textbf{0.412}   & 31.65         & \textbf{0.316}    & 27.64             & \textbf{0.521}    \\
0.75    & 37.59          & 0.432            & 22.70         & 0.366             & 20.59             & 0.557             \\
0.5     & 30.17          & 0.451            & 14.60         & 0.381             & 17.37             & 0.626             \\
0.25    & 23.27          & 0.481            & 13.65         & 0.410             & 12.53             & 0.653             \\
0       & \textbf{19.68} & 0.536            & \textbf{5.10} & 0.510             & \textbf{11.49}    & 0.679             \\
\hline
\end{tabular}
\caption{Quantitative comparison with different $\alpha$. %interpolation weights.
}
\vspace{-10mm}
\label{table:noise_blend}
\end{center}
\end{table}
\setlength{\tabcolsep}{0.5pt}

\begin{figure*}[t!]
\centering
\includegraphics[width=1\linewidth]{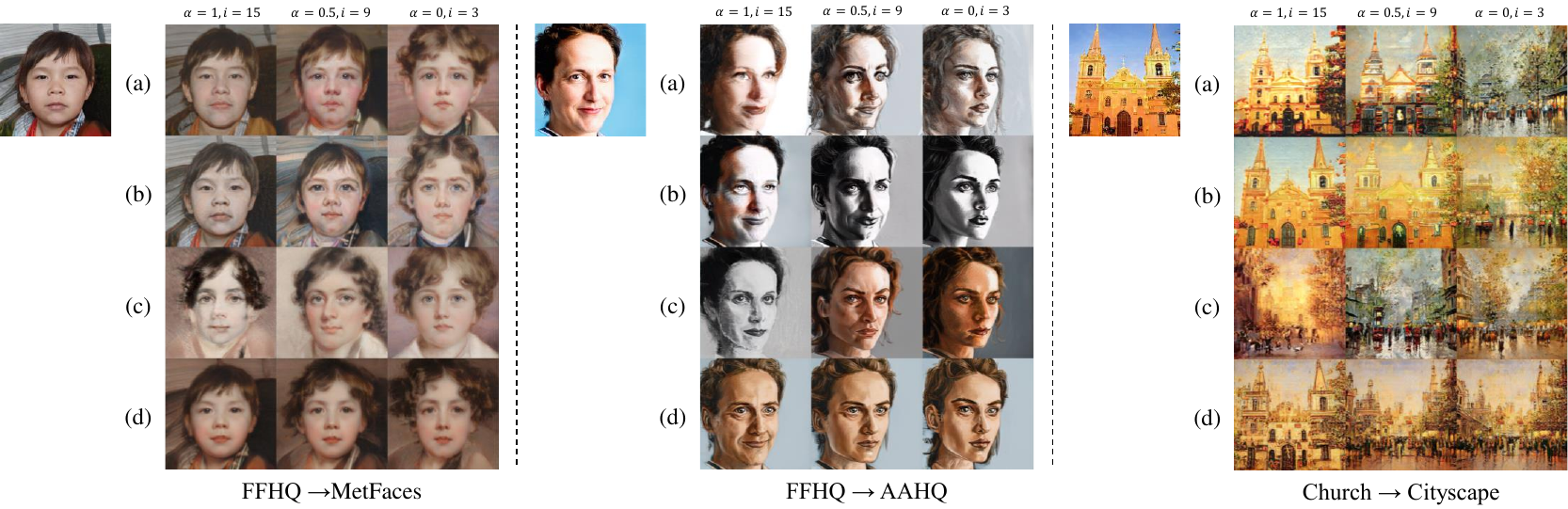}
   \vspace{-6mm}
   \caption{Qualitative comparison on controlling  preserved source features: (a) Layer-swap \cite{pinkney2020resolution}, (b) UI2I StyleGAN2 \cite{kwong2021unsupervised}, (c) Freeze G \cite{lee2020freezeg}, (d) ours. Our results shows consistent transition between the source and target features.}
   \vspace{-5mm}
\label{fig:compare_baseline}
\end{figure*}

\subsection{Effectiveness of our method}
\vspace{-1mm}
Fig.~\ref{fig:blending} presents our noise interpolation results on different settings. 
The source domain features are highly preserved in the images generated from the anchored subspace ($\alpha=1$), whereas they are lost in the rest of space ($\alpha=0$).
This indicates that FixNoise successfully disentangles the source and target features in $\mathcal H_{s \rightarrow t}$.
The fact that the features of both domains are embedded in a single space $\mathcal H_{s \rightarrow t}$ enables a smooth transition between the source and target features through interpolation between the anchor point and other points.
This also allows us to control the degree of preserved source features.
Further, we quantitatively examine the effects of the noise interpolation. FID \cite{heusel2017gans} and LPIPS \cite{zhang2018perceptual} are used to capture distance with the source and target distribution, respectively.
For LPIPS, we randomly sample 2000 latent codes $\mathbf z \in \mathcal Z$ and measure the distance between images generated by $G_s$ and $G_{s \rightarrow t}$ from the same $\mathbf z$.
A low FID indicates that the distribution of the generated images is close to the target data distribution, while low LPIPS indicates that the generated images are similar to the source images.
As shown in Table~\ref{table:noise_blend}, the generated images lose their source domain features and approach the target distribution as $\alpha$ decreases.

\setlength{\tabcolsep}{2.5pt}
\begin{table}[t!]
\begin{center}
\begin{tabular}{|c|rr|rr|rr|}
\hline
Source        & \multicolumn{4}{c|}{FFHQ}                                  & \multicolumn{2}{c|}{Church}        \\ \hline
Target        & \multicolumn{2}{c|}{MetFaces}  & \multicolumn{2}{c|}{AAHQ} & \multicolumn{2}{c|}{Cityscape} \\ \hline
  &  FID & \vtop{\hbox{KID} \hbox{\tiny{($\times 10^3)$}}}  & FID & \vtop{\hbox{KID} \hbox{\tiny{($\times 10^3)$}}}  &  FID & \vtop{\hbox{KID} \hbox{\tiny{($\times 10^3)$}}} \\
\hline
Layer-swap    &   68.31 & 34.69   &   38.03 & 28.30   &  52.02 & 38.46    \\
UI2I StyleGAN2    &   79.54 & 45.38   &   51.10 & 40.77   &  64.49 & 50.03    \\
Freeze G         &   24.12 & 5.41    &    7.93 &  2.82   &  12.84 & 3.55 \\
Ours ($\alpha = 1$)     &   40.37 & 14.88    &       31.65 & 22.86       &  27.64 & 16.28 \\
Ours ($\alpha = 0$)     &   \textbf{19.68} & \textbf{3.31}    &       \textbf{5.10} & \textbf{1.55}       &  \textbf{11.49} & \textbf{3.03} \\
\hline\hline
StyleGAN2-ADA        &   19.04 & 2.74    &    4.32 & 1.22    &  11.04 & \ \,2.75 \\
\hline
\end{tabular}
\caption{Quantitative comparison with baselines.}
\vspace{-10mm}
\label{table:2}
\end{center}
\end{table}
\setlength{\tabcolsep}{0.5pt}

\subsection{Comparison with baselines}
\vspace{-1mm}
We compare our approach with other methods that preserve source domain features during transfer learning including Freeze G \cite{lee2020freezeg}, Layer-swap \cite{pinkney2020resolution} and UI2I StyleGAN2 \cite{kwong2021unsupervised}.
For a qualitative comparison, we use the interpolation weight $\alpha = 1, 0.5, 0$ for ours, and freeze or swap layers as $i = 15, 9, 3$ for baselines to match a similar preservation level, respectively.
For a quantitative comparison, we freeze or swap layers as 15 for baselines following \cite{kwong2021unsupervised}.
When $i = 0$, $G_{s \rightarrow t}$ is fine-tuned without any constraints, and when $i = 21$, $G_{s \rightarrow t}$ is merely $G_s$.

The qualitative comparison of controlling preserved source features is shown in Fig.~\ref{fig:compare_baseline}. 
In similar domain settings, inconsistent transitions occur in competing methods.
Several unnatural color transitions are observed in Layer-swap and UI2I StyleGAN2 due to combining two different models.
Although the inconsistency and color transition problems are less important in Church $\rightarrow$ Cityscape, the feature control steps in the baselines are restricted to the number of layers, which interferes with a smooth transition.
In addition, to control the source features, previous methods require new models by swapping or training, which is not suitable for practical application.
In contrast, our method enables smooth transition through noise interpolation in a single model, which is easily applicable to diverse tasks.

The quantitative comparison is shown in Table~\ref{table:2}. When $\alpha = 0$, our method achieved similar performance with StyleGAN2-ADA which does not have any constraints on source preservation, while outperforming the other methods.
Furthermore, compared to Layer-swap and UI2I StylGAN2 that utilize weights of the source model, our method achieved a reasonable performance.
Although Freeze G obtained a better score than ours when $\alpha = 1$, our method shows more consistent results, as shown in Fig.~\ref{fig:compare_baseline}.

\section{Conclusion}
In this paper, we proposed a new transfer learning strategy, FixNoise, to control preserved source domain features in the target model.
FixNoise, combined with a simple feature matching loss, successfully disentangles the source and target features in the feature space of the target model.
Consequently, through the noise interpolation, our method can control the source features in a single model.
This demonstrates that our approach is superior to other methods in terms of computational cost.
Furthermore, experimental results reveal that the proposed method remarkably outperforms the previous works in terms of image quality and consistent transition.
We believe that our methods can be applied to various fields that utilize multi-domain features such as domain translation and image morphing.

%%%%%%%%% REFERENCES
{\small
\bibliographystyle{ieee_fullname}
\bibliography{egbib}
}

\ifarxiv
\clearpage
\appendix
\section{Training details}
\label{sec:training}
We build upon the base configuration in the official Pytorch \cite{paszke2017automatic} implementation of StyleGAN2-ADA\footnote{\scriptsize{\url{https://github.com/NVlabs/stylegan2-ada-pytorch}}} \cite{Karras2020ada}.
we train our model using $0.05$ loss weight for the proposed loss $\mathcal L_{fm}$ combined with other losses in StyleGAN2-ADA.
We used official pre-trained weights for source model trained on FFHQ\footnote{\scriptsize{\url{http://nvlabs-fi-cdn.nvidia.com/stylegan2-ada-pytorch/pretrained/transfer-learning-source-nets/ffhq-res256-mirror-paper256-noaug.pkl}}} \cite{karras2019style} and LSUN Church\footnote{\scriptsize{\url{http://nvlabs-fi-cdn.nvidia.com/stylegan2/networks/stylegan2-church-config-f.pkl}}} \cite{yu2015lsun}.
FFHQ $\rightarrow$ MetFaces \cite{Karras2020ada}, FFHQ $\rightarrow$ AAHQ \cite{liu2021blendgan} and LSUN Church $\rightarrow$ WikiArt Cityscape \cite{wikiart} are trained for 2000K, 12000K and 5000K images, respectively.
The batch size is set to 64.

\section{Ablation study on feature space}
\label{sec:ablation}
In FFHQ $\rightarrow$ MetFaces setting, we study in which space features are appropriate to preserve.
We conduct an experiment by applying a feature matching loss in intermediate feature space $\mathcal H$ (ours), RGB space, image space.
Fig.~\ref{fig:space} shows a qualitative comparison on applying the loss in the different spaces.
Results of the loss applied in the image and RGB space show well-preserved source features, however, they are not adapted to the target domain.
On the other hand, when the loss is applied on the intermediate feature space (ours), a target model $G_{s \rightarrow t}$ successfully learns features of the target domain while preserving the source features.
The feature matching loss in $\mathcal H$ allows tRGB layers to learn the target distribution.

\section{Additional results}
\label{sec:results}
\noindent\textbf{Noise interplation.} %\\ \indent
We provide additional noise interpolation results of the proposed method on FFHQ $\rightarrow$ MetFaces (Fig.~\ref{fig:ours_metface}), FFHQ $\rightarrow$ AAHQ (Fig.~\ref{fig:ours_aahq}) and LSUN Church $\rightarrow$ WikiArt Cityscape (Fig.~\ref{fig:ours_wikiart}).

\begin{figure}
    \centering
    \includegraphics[width=0.9\linewidth]{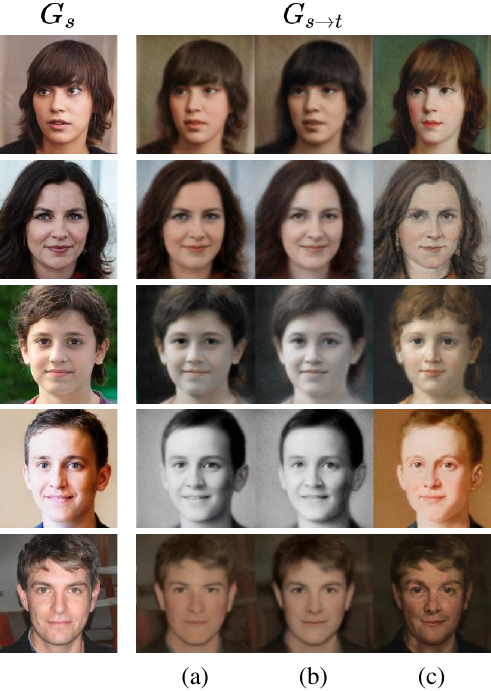}
    \caption{Visualizing the effects of the feature matching loss in different spaces: (a) image space, (b) RGB space, (c) intermediate feature space $\mathcal H$ (ours).}
    \label{fig:space}
\end{figure}

\noindent\textbf{Comparison with baselines.}
An additional cqualitative comparison of controlling preserved source features are shown in Fig. ~\ref{fig:compare_metface},~\ref{fig:compare_aahq} and ~\ref{fig:compare_wikiart}.
Freeze G \cite{lee2020freezeg} that requires a new training for each source degree shows an inconsistent transition of the preserved source features.
Layer-swap \cite{pinkney2020resolution} and UI2I StyleGAN2 \cite{kwong2021unsupervised} that convert weights of a source model also show the inconsistent transition.
Specifically, unnatural color transitions from the source domain are observed in Fig. ~\ref{fig:compare_metface}. Additionally, several artifacts and changes in the human identity are observed in Fig. ~\ref{fig:compare_aahq}.
We believe that this phenomenon occurs due to the long training time of the target model (\textit{e.g.} FFHQ $\rightarrow$ AAHQ are trained for 12000K images).
The long training time causes more changes in the target model weights, and this may disturb the combined models to generate realistic images.
For example, the identity changes seen in the result of layer swap (Fig.~\ref{fig:compare_aahq}) seems to be caused by a large change in the mapping function that transforms $\mathbf z \in \mathcal Z$ to $\mathbf w \in \mathcal W$.
The color transition problems and inconsistent transition are less observable in LSUN Church $\rightarrow$ WikiArt Cityscape, due to the artistic target dataset and the spatial difference between the source and target domain, respectively.
Nevertheless, these methods require models for each degree of preserved source features, while  the proposed method can control in a single model.

\noindent\textbf{Latent modulation.}
Recently, several works \cite{harkonen2020ganspace,shen2021closed,wu2021stylespace} observe that StyleGAN can effectively adjust semantic attributes of images by modulating latent codes in interpretable directions.
Additionally, StyleSpace \cite{wu2021stylespace} revealed that the $\mathcal{S}$ space is the most disentangled among the three latent spaces $\mathcal{Z}$, $\mathcal{W}$, and $\mathcal{S}$ of StyleGAN \cite{karras2019style,karras2020analyzing}, and it is possible to change various semantic attributes of generated images just by adjusting a value of single dimension of $\mathcal S$.
Based on this observation, we examine latent modulation effects on our proposed method. The latent modulation effects on different interpolation weights are shown in Fig.~\ref{fig:latent_metface},\ref{fig:latent_aahq} and \ref{fig:latent_wikiart}.
The latent modulation effects of the source model are highly aligned in anchored subspace ($\alpha = 1$).
As $\alpha$ decreases, some latent modulation effects remain, while the rest gradually weakens or disappears.
This phenomenon may occur as the preserved source features gradually vanish.

\begin{figure*}[t!]
\centering
\includegraphics[width=0.9\linewidth]{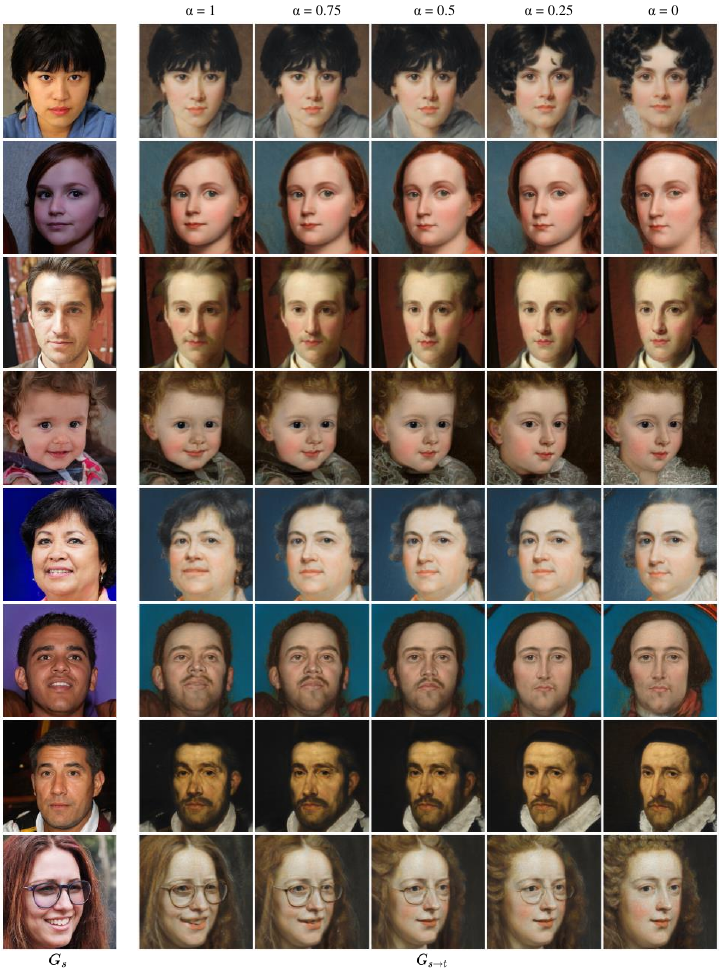}
   \caption{\textbf{[FFHQ $\rightarrow$ MetFaces]}
   Visualizing the effects of the noise interpolation.
   The interpolation weight $\alpha$ is presented above each column.}
\label{fig:ours_metface}
\end{figure*}

\begin{figure*}[t!]
\centering
\includegraphics[width=0.9\linewidth]{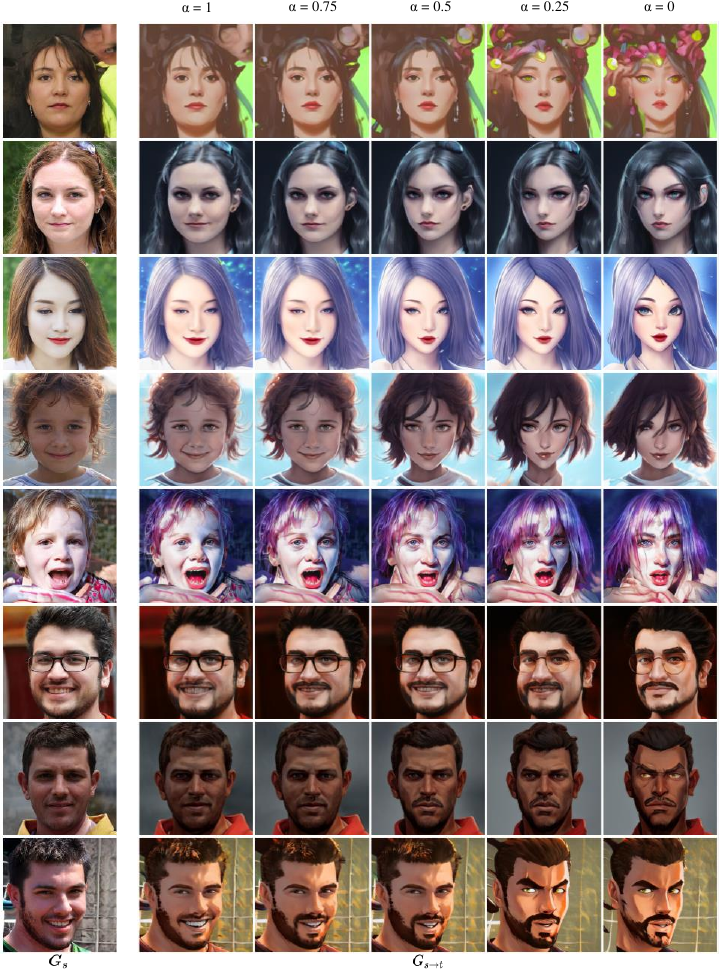}
   \caption{\textbf{[FFHQ $\rightarrow$ AAHQ]}
   Visualizing the effects of the noise interpolation.
   The interpolation weight $\alpha$ is presented above each column.}
\label{fig:ours_aahq}
\end{figure*}

\begin{figure*}[t!]
\centering
\includegraphics[width=0.9\linewidth]{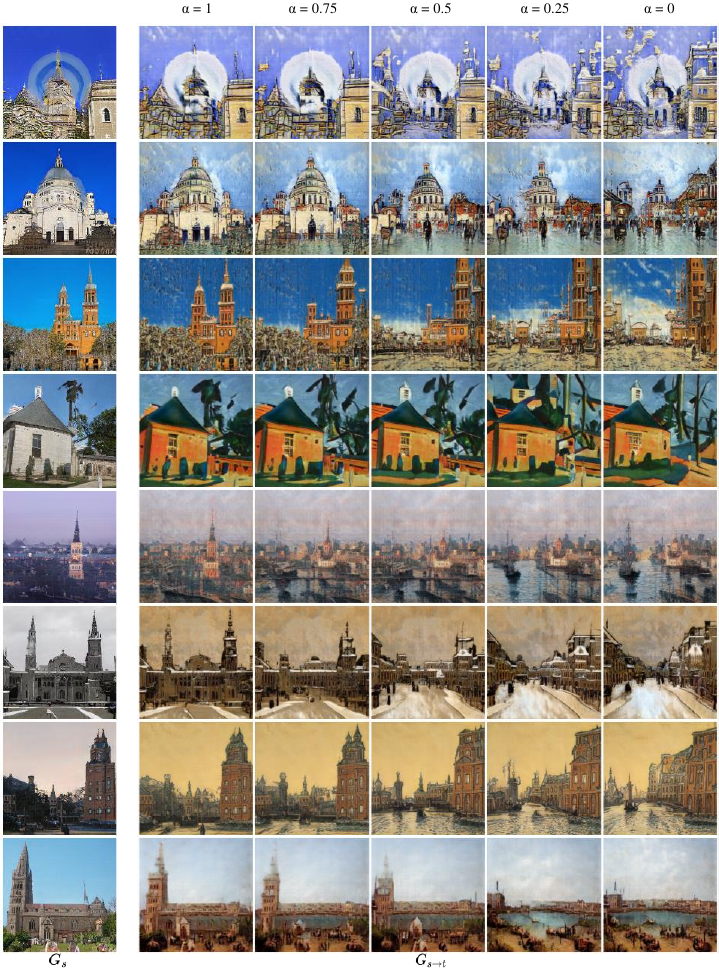}
   \caption{\textbf{[Church $\rightarrow$ Cityscape]}
   Visualizing the effects of the noise interpolation.
   The interpolation weight $\alpha$ is presented above each column.}
\label{fig:ours_wikiart}
\end{figure*}

\begin{figure*}[t!]
\centering
\includegraphics[width=0.9\linewidth]{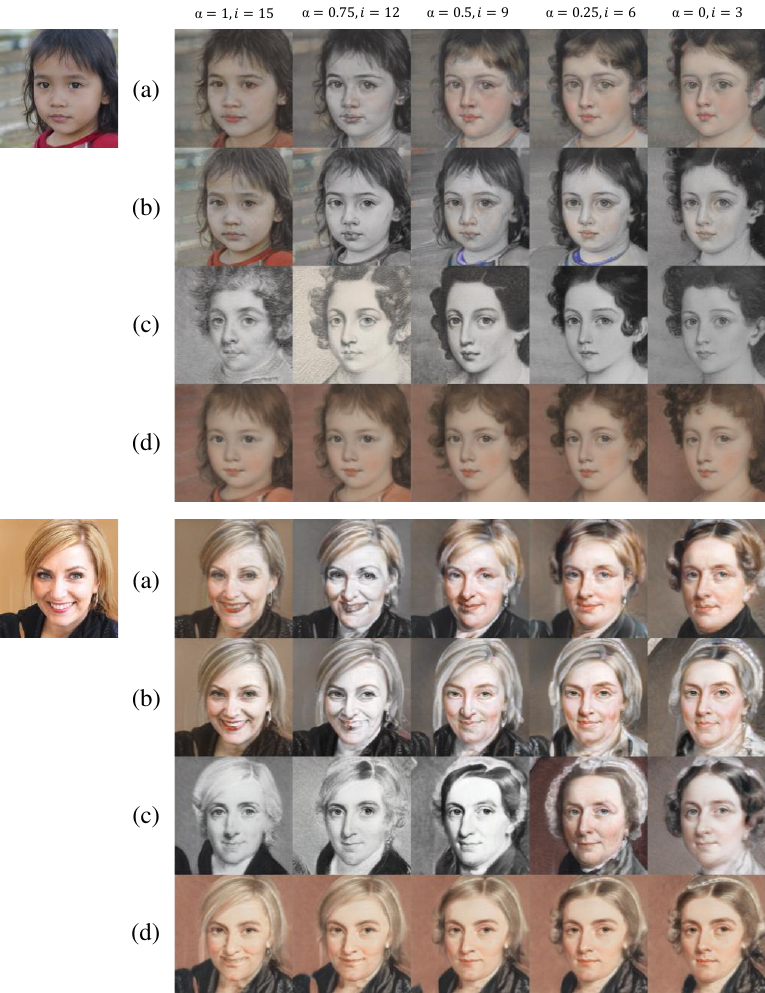}
   \caption{\textbf{[FFHQ $\rightarrow$ MetFaces]} Qualitative comparison on controlling  preserved source features: (a) Layer-swap \cite{pinkney2020resolution}, (b) UI2I StyleGAN2 \cite{kwong2021unsupervised}, (c) Freeze G \cite{lee2020freezeg}, (d) ours.
   The interpolation weight $\alpha$ and swap / freeze layer $i$ are presented above each column.}
\label{fig:compare_metface}
\end{figure*}

\begin{figure*}[t!]
\centering
\includegraphics[width=0.9\linewidth]{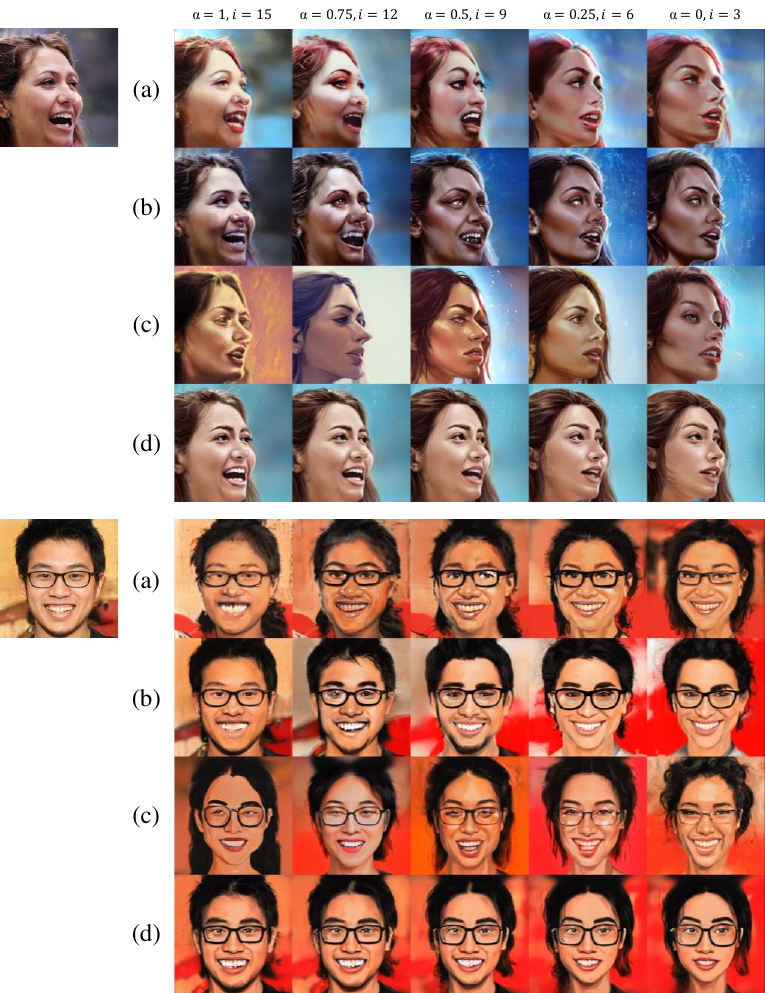}
   \caption{\textbf{[FFHQ $\rightarrow$ AAHQ]} Qualitative comparison on controlling  preserved source features: (a) Layer-swap \cite{pinkney2020resolution}, (b) UI2I StyleGAN2 \cite{kwong2021unsupervised}, (c) Freeze G \cite{lee2020freezeg}, (d) ours.
   The interpolation weight $\alpha$ and swap / freeze layer $i$ are presented above each column.}
\label{fig:compare_aahq}
\end{figure*}

\begin{figure*}[t!]
\centering
\includegraphics[width=0.9\linewidth]{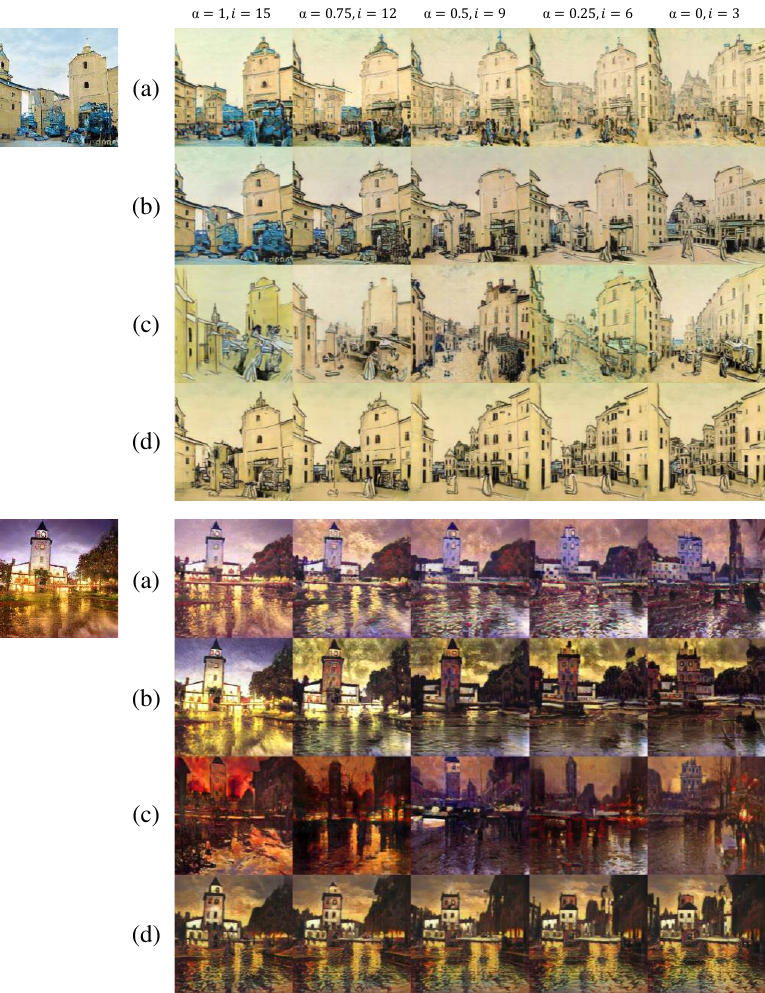}
   \caption{\textbf{[Church $\rightarrow$ Cityscape]} Qualitative comparison on controlling  preserved source features: (a) Layer-swap \cite{pinkney2020resolution}, (b) UI2I StyleGAN2 \cite{kwong2021unsupervised}, (c) Freeze G \cite{lee2020freezeg}, (d) ours.
   The interpolation weight $\alpha$ and swap / freeze layer $i$ are presented above each column.}
\label{fig:compare_wikiart}
\end{figure*}

\begin{figure*}[t!]
\centering
\includegraphics[width=0.9\linewidth]{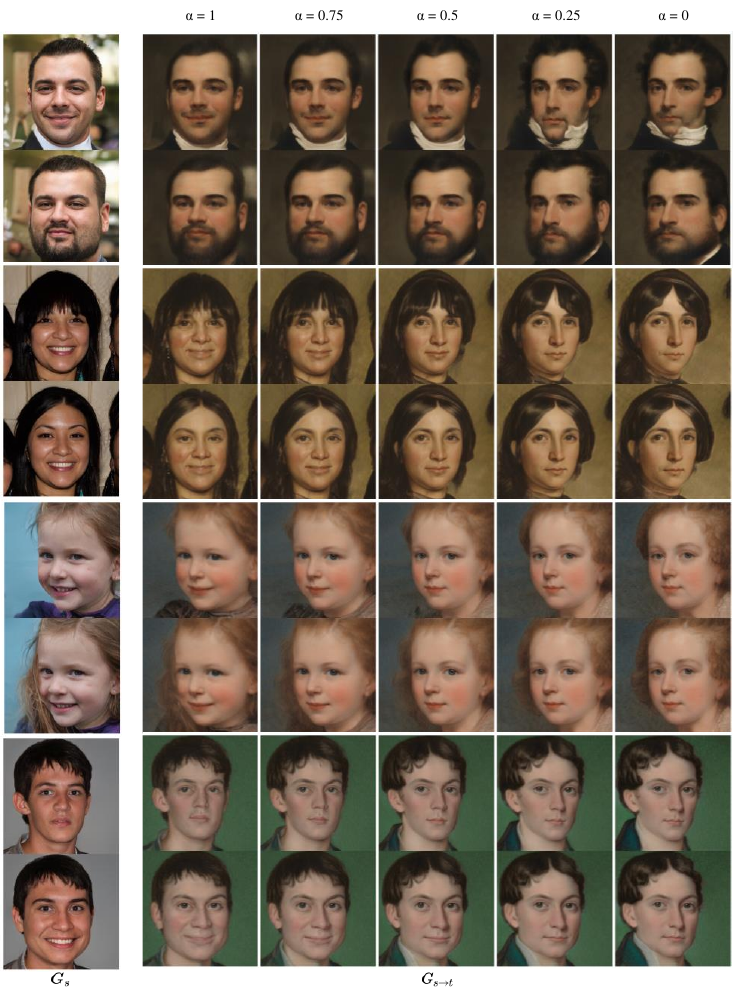}
   \caption{\textbf{[FFHQ $\rightarrow$ MetFaces]}
   Visualizing the effects of the latent modulation on different interpolation weight.
   Each of the two adjacent columns is the result of modulating the latent in a different direction ($+/-$).
   The interpolation weight $\alpha$ is presented above each column.}
\label{fig:latent_metface}
\end{figure*}

\begin{figure*}[t!]
\centering
\includegraphics[width=0.9\linewidth]{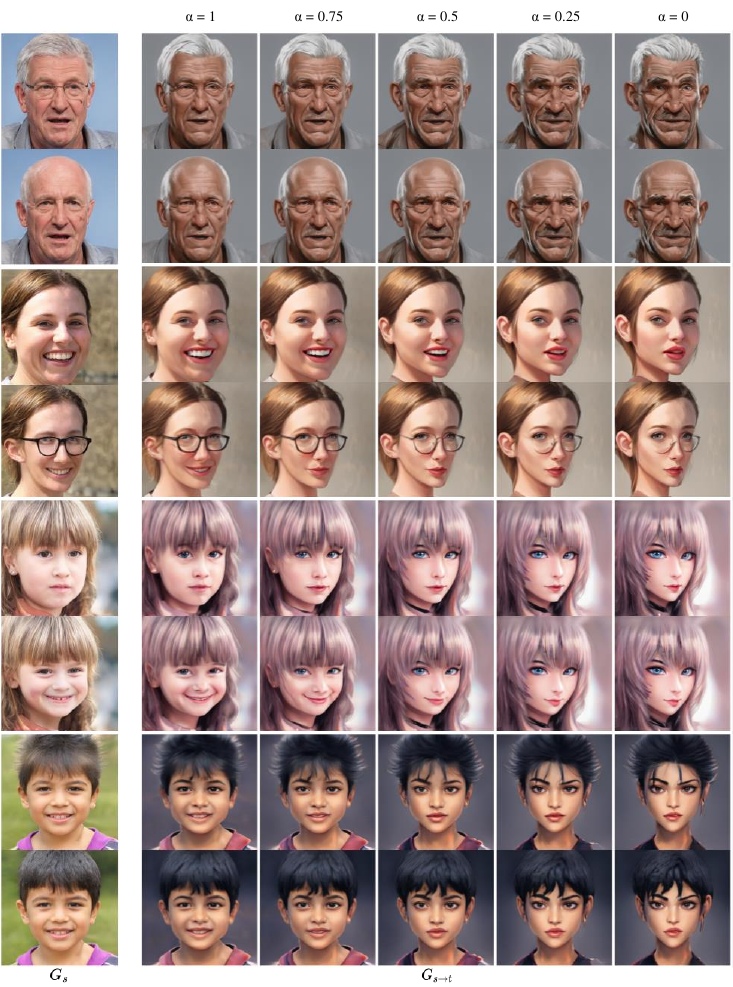}
   \caption{\textbf{[FFHQ $\rightarrow$ AAHQ]}
   Visualizing the effects of the latent modulation on different interpolation weight.
   Each of the two adjacent columns is the result of modulating the latent in a different direction ($+/-$).
   The interpolation weight $\alpha$ is presented above each column.}
\label{fig:latent_aahq}
\end{figure*}

\begin{figure*}[t!]
\centering
\includegraphics[width=0.9\linewidth]{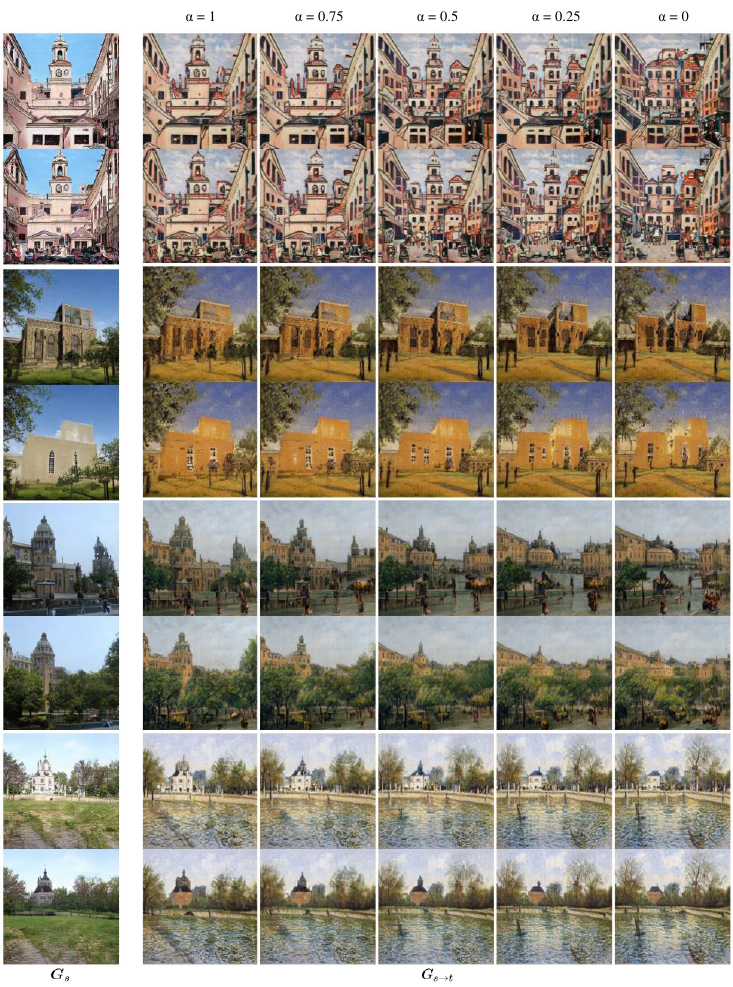}
   \caption{\textbf{[Church $\rightarrow$ Cityscape]}
   Visualizing the effects of the latent modulation on different interpolation weight.
   Each of the two adjacent columns is the result of modulating the latent in a different direction ($+/-$).
   The interpolation weight $\alpha$ is presented above each column.}
\label{fig:latent_wikiart}
\end{figure*}
\fi

\end{document}